\documentclass[11pt,letterpaper]{article}
\usepackage{emnlp2016}
\usepackage{times}
\usepackage{latexsym}
\usepackage{graphicx}

\emnlpfinalcopy

\def\emnlppaperid{***}

\title{Identifying Dogmatism in Social Media: Signals and Models}

\author{Ethan Fast \and Eric Horvitz\\
  {\tt ethaen@stanford.edu, horvitz@microsoft.com}}

\date{}

\begin{document}

\maketitle

\begin{abstract}
We explore linguistic and behavioral features of dogmatism in social media and construct statistical models that can identify dogmatic comments. Our model is based on a corpus of Reddit posts, collected across a diverse set of conversational topics and annotated via paid crowdsourcing. We operationalize key aspects of dogmatism described by existing psychology theories (such as over-confidence), finding they have predictive power. We also find evidence for new signals of dogmatism, such as the tendency of dogmatic posts to refrain from signaling cognitive processes. When we use our predictive model to analyze millions of other Reddit posts, we find evidence that suggests dogmatism is a deeper personality trait, present for dogmatic users across many different domains, and that users who engage on dogmatic comments tend to show increases in dogmatic posts themselves.
\end{abstract}

\section{Introduction}

\footnotesize
\begin{quote}
``I'm supposed to trust the opinion of a MS minion? The people that produced Windows ME, Vista and 8? They don't even understand people, yet they think they can predict the behavior of new, self-guiding AI?'' --\textit{anonymous} 
\end{quote}
\begin{quote}
``I think an AI
would make it easier for Patients to confide their information because by nature, a robot cannot judge them. Win-win? :D''' --\textit{anonymous}
\end{quote}

\normalsize

Dogmatism describes the tendency to lay down opinions as incontrovertibly true, without respect for conflicting evidence or the opinions of others \cite{dogma-definition}. Which user is more dogmatic in the examples above? This question is simple for humans. Phrases like ``they think'' and ``they don't even understand,'' suggest an intractability of opinion, while ``I think'' and ``win-win?'' suggest the opposite. Can we train computers to draw similar distinctions? Work in psychology has called out many aspects of dogmatism that can be modeled computationally via natural language, such as over-confidence and strong emotions \cite{dogmatism}.



We present a statistical model of dogmatism that addresses two complementary goals. First, we validate psychological theories by examining the predictive power of feature sets that guide the model's predictions. For example, do linguistic signals of certainty help to predict a post is dogmatic, as theory would suggest? Second, we apply our model to answer four questions:

\textbf{R1}: What kinds of topics (e.g., guns, LGBT) attract the highest levels of dogmatism?

\textbf{R2}: How do dogmatic beliefs cluster? 

\textbf{R3}: How does dogmatism influence a conversation on social media? 

\textbf{R4}: How do other user behaviors (e.g., frequency and breadth of posts) relate to dogmatism?

We train a predictive model to classify dogmatic posts from Reddit, one of the most popular discussion communities on the web.\footnote{http://www.reddit.com} Posts on Reddit capture discussion and debate across a diverse set of domains and topics -- users talk about everything from climate change and abortion, to world news and relationship advice, to the future of artificial intelligence. 
As a prerequisite to training our model, we have created a corpus of 5,000 Reddit posts annotated with levels of dogmatism, which we are releasing to share with other researchers.

Using the model, we operationalize key domain-independent aspects of psychological theories of dogmatism drawn from the literature. We find these features have predictive power that largely supports the underlying theory. For example, posts that use less confident language tend to be less dogmatic. We also discover evidence for new attributes of dogmatism. For example, dogmatic posts tend not to verbalize cognition, through terms such as ``I think,'' ``possibly,'' or ``might be.''

Our model is trained on only 5,000 annotated posts, but once trained, we use it to analyze millions of other Reddit posts to answer our research questions. We find a diverse set of topics are colored by dogmatic language (e.g., people are dogmatic about religion, but also about LGBT issues). Further, we find some evidence for dogmatism as a deeper personality trait -- people who are strongly dogmatic about one topic are more likely to express dogmatic views about others as well. Finally, in conversation, we discover that one user's dogmatism tends to bring out dogmatism in their conversational partner, forming a vicious cycle.  

\begin{figure}[t!]
\includegraphics[width=1.0\columnwidth]{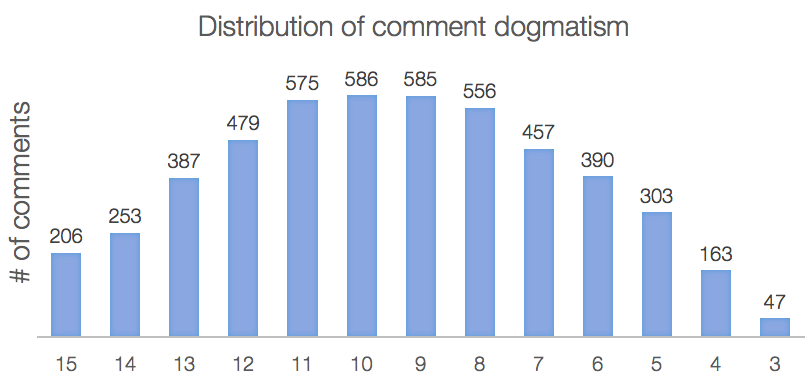}
\caption{We crowdsourced dogmatism labels for 5000 comments. The distribution is slightly skewed towards higher levels of dogmatism. For example, crowdworkers unanimously labeled 206 comments as highly dogmatic ($5\times 3=15$), but only 47 as minimally dogmatic ($1\times 3=3$).}
\label{fig:dogma-comments}
\end{figure}

\section{Dogmatism data}

Posts on Reddit capture debate and discussion across a diverse set of topics, making them a natural starting point for untangling domain-independent linguistic features of dogmatism.

\textbf{Data collection.} Subreddits are sub-communities on Reddit oriented around specific interests or topics, such as \textit{technology} or \textit{politics}. Sampling from Reddit as a whole would bias the model towards the most commonly discussed content. But by sampling posts from individual subreddits, we can control the kinds of posts we use to train our model. To collect a diverse training dataset, we have randomly sampled 1000 posts each from the subreddits \textit{politics}, \textit{business}, \textit{science}, and \textit{AskReddit}, and 1000 additional posts from the Reddit frontpage. All posts in our sample appeared between January 2007 and March 2015, and to control for length effects, contain between 300 and 400 characters. This results in a total training dataset of 5000 posts.

\textbf{Dogmatism annotations.} Building a useful computational model requires labeled training data. We labeled the Reddit dataset using crowdworkers on Amazon Mechanical Turk (AMT), creating the first public corpus annotated with levels of dogmatism. We asked crowdworkers to rate levels of dogmatism on a 5-point Likert scale, as supported by similar annotation tasks in prior work \cite{politeness}. Concretely, we gave crowdworkers the following task:
\small
\begin{quote}
Given a comment, imagine you hold a well-informed, different opinion from the commenter in question. We'd like you to tell us how likely that commenter would be to engage you in a constructive conversation about your disagreement, where you each are able to explore the other's beliefs. The options are:

\textbf{(5)}: It's unlikely you'll be able to engage in any substantive conversation. When you respectfully express your disagreement, they are likely to ignore you or insult you or otherwise lower the level of discourse.

\textbf{(4)}: They are deeply rooted in their opinion, but you are able to exchange your views without the conversation degenerating too much.

\textbf{(3)}: It's not likely you'll be able to change their mind, but you're easily able to talk and  understand each other's point of view.

\textbf{(2)}: They may have a clear opinion about the subject, but would likely be open to discussing alternative viewpoints.

\textbf{(1)}: They are not set in their opinion, and it's possible you might change their mind. If the comment does not convey an opinion of any kind, you may also select this option.

\end{quote}
\normalsize
To ensure quality work, we restricted the task to Masters workers and provided examples corresponding to each point on the scale. Including examples in a task has been shown to significantly increase the agreement and quality of crowdwork \cite{crowd-examples}. For instance, here is an example of a highly dogmatic \textit{(5)} comment:
\small
\begin{quote}
I won't be happy until I see the executive suite of BofA, Wells, and all the others, frog-marched into waiting squad cars. It's ALREADY BEEN ESTABLISHED that...
\end{quote}
\normalsize
And a minimally dogmatic \textit{(1)} comment:
\small
\begin{quote}
I agree. I would like to compile a playlist for us trance yogi's, even if you just would like to experiment with it. Is there any preference on which platform to use?
\end{quote}
\normalsize
Each comment has been annotated by three independent workers on AMT, which is enough to produce reliable results in most labeling tasks \cite{get-another-label}. To compute an aggregate measure of dogmatism for each comment, we summed the scores of all three workers. We show the resulting distribution of annotations in Figure \ref{fig:dogma-comments}.

\textbf{Inter-annotator agreement.} To evaluate the reliability of annotations we compute Krippendorff's $\alpha$, a measure of agreement designed for variable levels of measurement such as a Likert scale \cite{krippendorff}. An $\alpha$ of $0$ indicates agreement indistinguishable from chance, while an $\alpha$ of 1 indicates perfect agreement. Across all annotations we find $\alpha=0.44$. While workers agree much more than chance, clearly dogmatism is also subjective. In fact, when we examine only the middle two quartiles of the dogmatism annotations, we find agreement is no better than chance. Alternatively, when we measure agreement only among the top and bottom quartiles of annotations, we find agreement of $\alpha=0.69$. This suggests comments with scores that are only slightly dogmatic are unreliable and often subject to human disagreement. For this reason, we use only the top and bottom quartiles of comments when training our model.

\begin{table*}[!ht]
  \renewcommand{\arraystretch}{1.2}
  \begin{tabular}{p{3.5cm}p{1.5cm}p{10cm}}
    \textbf{Strategy} & \textbf{Odds} & \textbf{Example} \\
    \hline
    Certainty & 1.33* & Be a hate monger \textbf{all} you want... Your life will \textbf{never} truly be happy though, and you will \textbf{never} know peace. \\
    Tentativeness & 0.88* & Most are \textbf{likely} to be more technically advanced and, \textbf{if} still using radio, \textbf{might} very well be emitting signals we could detect \\
    Insight & 0.83* & I \textbf{think} stating the obvious is a necessary function. Information like this is important to \textbf{consider}... \\
    Perception & 0.77* &  I \textbf{saw} four crows on that same branch, \textbf{staring} at the deceased. The \textbf{silence} of the crows was \textbf{deafening}. \\
    Relativity & 0.82* & I've known a number to go \textbf{into} shock \textbf{during} the procedure \\
    Comparison & 0.91 & This may be \textbf{more} \textbf{than} a coincidence. \\
    I (pronouns) & 0.68* & Like \textbf{I} said, \textbf{I} want to believe the former. \textbf{I'm} glad it worked out.\\
    You (pronouns) & 2.18* & I don't give a fuck what \textbf{you} do. \textbf{You} can get drink \textbf{yourself} to death, \textbf{you} can get \textbf{yourself} pregnant... \\
    We (pronouns) & 0.96 & \textbf{We} need a bigger, better, colder fridge. \textbf{We} have worked hard... \\
     They (pronouns) & 1.63* & \textbf{They} want the ability to prosecute who they please. \\
    Past & 0.69* &  I \textbf{was} walking past and \textbf{thought} about asking if they \textbf{needed} help. \\
    Present & 1.11* & Can I \textbf{steal} your organs and nutrients if I \textbf{need} them and you don't \textbf{want} to give them up? \\
    Future & 1.06 & Trump's thugs \textbf{will} be pretending to be Bernie supporters and \textbf{will} set fire to Philadelphia. \\
    Interrogatory & 1.12* & Gee, \textbf{where} was the NY Times back in the day? \textbf{Why} didn't we hear of the Kennedys, LBJ and FDR? \\
    Negation & 1.35* & If you \textbf{didn't} know the woman well enough to know she \textbf{didn't} take BC regularly, you certainly \textbf{don't} know her well enough to know she \textbf{doesn't} have an std.\\
    Negative emotion & 2.32* & A prank?!? You \textbf{arrogant} son of a \textbf{bitch} \\ 
    Positive emotion & 0.96 & They were \textbf{excellent} fishermen - they built \textbf{fine} boats. \\
  \end{tabular}
  \caption{Linguistic features that capture high level psychological categories and their relationship with dogmatic comments. \textit{Strategy} describes the psychological category. \textit{Odds} describes the likelihood that a category will appear more often in a dogmatic comment (e.g., dogmatic comments are 2.18 times more likely to mention \textit{you}-oriented phrases). \textit{Example} illustrates a comment that matches the category. * indicates significance ($p<0.05$) after correction with Holmes method.}
  \label{tbl:features}
\end{table*}

\section{Approaches to Identifying Dogmatism}

We now consider strategies for identifying dogmatism based on prior work in psychology. We start with the Linguistic Inquiry and Word Count (LIWC), a lexicon popular in the social sciences \cite{liwc}. LIWC provides human validated lists of words that correspond to high-level psychological categories such as \textit{certainty} or \textit{perception}. In other studies, LIWC has uncovered linguistic signals relating to politeness \cite{politeness}, deception \cite{deception}, or authority in texts \cite{enron}. Here, we examine how dogmatism relates to 17 of LIWC's categories (Table \ref{tbl:features}).

To compute the relationships between LIWC categories and dogmatism, we first count the relevant category terms that appear in each annotated Reddit comment, normalized by its word count. We then calculate odds ratios on the aggregate counts of each LIWC category over the top and bottom quartiles of dogmatic comments. As we have discussed, using the top and bottom quartiles of comments provides a more reliable signal of dogmatism. We check for significant differences in categories between dogmatic and non-dogmatic comments using the Mann-Whitney U test and apply Holmes method for correction. All odds we report in this section are significant after correction.

Dogmatic statements tend to express a high degree of certainty \cite{dogmatism}. Here we consider LIWC categories that express certainty both positively (\textit{certainty}) and negatively (\textit{tentativeness}). For example, the word ``always'' is  certain, while ``possibly'' is tentative. Conforming to existing theory, \textit{certainty} is more associated with dogmatic comments (1.52 odds), while \textit{tentativeness} is more associated with the absence of dogmatism (0.88 odds).

Terms used to verbalize cognition can act as a hedge that often characterizes non-dogmatic language. LIWC's \textit{insight} category captures this effect through words such as ``think,'' ``know,'' or ``believe.'' These words add nuance to a statement \cite{cogproc-insight}, signaling it is the product of someone's mind (``\textit{I think} you should give this paper a good review'') and not meant to be interpreted as an objective truth. Along these lines, we find the use of terms in the \textit{insight} category is associated with non-dogmatic comments (0.83 odds).

Sensory language, with its focus on description and detail, often signals a lack of any kind of opinion, dogmatic or otherwise. LIWC's \textit{perception} category captures this idea through words associated with hearing, feeling, or seeing. For example, these words might occur when recounting a personal experience (``I saw his incoming fist''), which even if emotionally charged or negative, is less likely to be dogmatic. We find perception is associated with non-dogmatic comments at 0.77 odds. 

Drawing comparisons or qualifying something as relative to something else conveys a nuance that is absent from traditionally dogmatic language. The LIWC categories \textit{comparison} and \textit{relativity} capture these effects through comparison words such as ``than'' or ``as'' and qualifying words such as ``during'' or ``when.'' For example, the statement ``I hate politicians'' is more dogmatic than ``I hate politicians \textit{when} they can't get anything done.' \textit{Relativity} is associated with non-dogmatic comments at 0.80 odds, but \textit{comparison} does not reach significance. 

Pronouns can be surprisingly revealing indicators of language: for example, signaling one's gender or hierarchical status in a conversation \cite{secret-life-pronouns}. We find first person singular pronouns are a useful negative signal for dogmatism (0.46 odds), while second person singular pronouns (2.18 odds) and third person plural (1.63 odds) are a useful positive signal. Looking across the corpus, we see \textit{I} often used with a hedge (``I think'' or ``I know''), while \textit{you} and \textit{they} tend to characterize the beliefs of others, often in a strongly opinionated way (``you are a moron'' or ``they are keeping us down''). Other pronoun types do not show significant relationships.


Like pronouns, verb tense can reveal subtle signals in language use, such as the tendency of medical inpatients to focus on the past \cite{ed-liwc}. On social media, comments written in the \textit{present tense} are more likely to be oriented towards a user's current interaction (``this \textit{is} all so stupid''), creating opportunities to signal dogmatism. Alternatively, comments in the \textit{past tense} are more likely to refer to outside experiences (``it \textit{was} an awful party''), speaking less to a user's stance towards an ongoing discussion. We find \textit{present tense} is a positive signal for dogmatism (1.11 odds) and \textit{past tense} is a negative signal (0.69 odds).

Dogmatic language can be either positively or negatively charged in sentiment: for example, consider the positive statement ``\textit{Trump is the SAVIOR of this country!!!}'' or the negative statement ``\textit{Are you REALLY that stupid?? Education is the only way out of this horrible mess. It's hard to imagine how anyone could be so deluded.}'' In diverse communities, where people hold many different kinds of opinions, dogmatic opinions will often tend to come into conflict with one another \cite{community-conflict}, producing a greater likelihood of negative sentiment. Perhaps for this reason, \textit{negative emotion} (2.09 odds) and \textit{swearing} (3.80 odds) are useful positive signals of dogmatism, while \textit{positive emotion} shows no significant relationship.

Finally, we find that \textit{interrogative} language (1.12 odds) and \textit{negation} (1.35 odds) are two additional positive signals of dogmatism. While interrogative words like ``how'' or ``what'' have many benign uses, they disproportionately appear in our data in the form of rhetorical or emotionally charged questions, such as ``how can anyone be that dumb?'' 

Many of these linguistic signals are correlated with each other, suggesting that dogmatism is the cumulative effect of many component relationships. For example, consider the relatively non-dogmatic statement: ``I think the reviewers are wrong in this instance.'' Removing signals of \textit{insight}, we have: ``the reviewers are wrong in this instance,'' which is slightly more dogmatic. Then removing \textit{relativity}, we have: ``the reviewers are wrong.'' And finally, adding \textit{certainty}, we have a dogmatic statement: ``the reviewers are always wrong.''

\begin{table}[tb]
\renewcommand{\arraystretch}{1.2}
  \begin{tabular}{p{7em}@{\hspace{2em}}p{5em}@{\hspace{1em}}p{6em}}
  \textbf{Classifier} & \textbf{In-domain} & \textbf{Cross-domain} \\
  \hline
BOW & 0.853 & 0.776 \\
SENT & 0.677 & 0.646 \\
LING & 0.801 & 0.728 \\
BOW + SENT & 0.860 & 0.783 \\
BOW + LING & 0.881 & 0.791  \\
\end{tabular}
\caption{The AUC scores for dogmatism classifiers within and across domains. BOW (bag-of-words) and SENT (sentiment signals) are baselines, and LING uses the linguistic features from Table 1. We compute in-domain accuracy using 15-fold cross-validation on the Reddit dataset, and cross-domain accuracy by training on Reddit and evaluating on comments on articles from the New York Times. Chance AUC is 0.5.}
  \label{tbl:classification}
\end{table}

\section{Predicting dogmatism}

We now show how we can use the linguistic feature sets we have described to build a classifier that predicts dogmatism in comments. A predictive model further validates our feature sets, and also allows us to analyze dogmatism in millions of other Reddit comments in a scalable way, with multiple uses in ongoing, downstream analyses. 

\textbf{Prediction task.} Our goal is (1) to understand how well we can use the strategies in Section 3 to predict dogmatism, and (2) to test the \textit{domain-independence} of these strategies. First, we test the performance of our model under cross-validation within the Reddit comment dataset. We then evaluate the Reddit-based model on a held out corpus of New York Times comments annotated using the technique in Section 2. We did not refer to this second dataset during feature construction.

For classification, we consider two classes of comments: \textit{dogmatic} and \textit{non-dogmatic}. As in the prior analysis, we draw these comments from the top and bottom quartiles of the dogmatism distribution. This means the classes are balanced, with 2,500 total comments in the Reddit training data and 500 total comments in the New York Times testing data.

We compare the predictions of logistic regression models based on unigram bag-of-words features (BOW),  sentiment signals\footnote{For SENT, we use normalized word counts from LIWC's positive and negative emotional categories.} (SENT), the linguistic features from our earlier analyses (LING), and combinations of these features. BOW and SENT provide baselines for the task. We compute BOW features using term frequency-inverse document frequency (TF-IDF) and category-based features by normalizing counts for each category by the number of words in each document. The BOW classifiers are trained with regularization (L2 penalties of 1.5).

\textbf{Classification results.} We present classification accuracy in Table \ref{tbl:classification}. BOW shows an AUC of 0.853 within Reddit and 0.776 on the held out New York Times comments. The linguistic features boost classification results within Reddit (0.881) and on the held out New York Times comments (0.791). While linguistic signals by themselves provide strong predictive power (0.801 AUC within domain), sentiment signals are much less predictive.

These results suggest that linguistic features inspired by prior efforts in psychology are useful for predicting dogmatism in practice and generalize across new domains.

\section{Dogmatism in the Reddit Community }

We now apply our dogmatism classifier to a larger dataset of posts, examining how dogmatic language shapes the Reddit community. Concretely, we apply the BOW+LING model trained on the full Reddit dataset to millions of new unannotated posts, labeling these posts with a probability of dogmatism according to the classifier (0=non-dogmatic, 1=dogmatic). We then use these dogmatism annotations to address four research questions.

\begin{table}[tb]
\renewcommand{\arraystretch}{1.2}
  \small
  \begin{tabular}{p{6em}@{\hspace{2.5em}}p{3em}@{\hspace{.5em}}p{6em}@{\hspace{1.25em}}p{2.5em}}
  \textbf{Highest} & \textbf{Score} & \textbf{Lowest} & \textbf{Score} \\
  \hline
cringepics & 0.553 & photography & 0.399 \\
DebateAChristian & 0.551 & DIY & 0.399 \\
DebateReligion & 0.540 & homebrewing & 0.401 \\
politics & 0.536 & cigars & 0.402 \\
ukpolitics & 0.533 & wicked\_edge & 0.404 \\
atheism & 0.529 & guitar & 0.406 \\
lgbt & 0.527 & gamedeals & 0.406 \\
TumblrInAction & 0.524 & buildapc & 0.407 \\
islam & 0.523 & techsupport & 0.410 \\
SubredditDrama & 0.520 & travel & 0.410 \\
\end{tabular}
\caption{Subreddits with the highest and lowest dogmatism scores. Politics and religion are common themes among the most dogmatic subreddits, while hobbies (e.g., photography, homebrewing, buildapc) show the least dogmatism.}
  \label{tbl:subreddits}
\end{table}
\normalsize

\subsection{What subreddits have the highest and lowest levels of dogmatism? (R1)}

A natural starting point for analyzing dogmatism on Reddit is to examine how it characterizes the site's sub-communities. For example, we might expect to see that subreddits oriented around topics such as abortion or climate change are more dogmatic, and subreddits about cooking are less so.

To answer this question, we randomly sample 1.6 million posts from the entire Reddit community between 2007 and 2015. We then annotate each of these posts with dogmatism using our classifier, and compute the average dogmatism level for each subreddit in the sample with at least 100 posts.

We present the results of this analysis in Table \ref{tbl:subreddits}. The subreddits with the highest levels of dogmatism tend to be oriented around politics and religion (\textit{DebateAChristian} or \textit{ukpolitics}), while those with the lowest levels tend to focus on hobbies (\textit{photography} or \textit{homebrewing}). The subreddit with the highest average dogmatism level, \textit{cringepics}, is a place to make fun of socially awkward messages, often from would-be romantic partners. Dogmatism here tends to take the form of ``how could someone be that stupid'' and is directed at the subject of the post, as opposed to other members of the community.

Similarly, \textit{SubredditDrama} is a community where people come to talk about fights on the internet or social media. These fights are often then extended in discussion, for example: \textit{``If the best you can come up with is that something you did was legal, it's probably time to own up to being an ass.''} The presence of this subreddit in our analysis provides a further sanity check that our model is capturing a robust signal of dogmatism.

\begin{table*}[tb]\scriptsize
\renewcommand{\arraystretch}{1.1}
  \begin{tabular}{p{6em}@{\hspace{4em}}p{6em}@{\hspace{4em}}p{6em}@{\hspace{4em}}p{6em}@{\hspace{4em}}p{6em}@{\hspace{4em}}p{6em}@{\hspace{4em}}p{5em}}
  \textbf{Libertarianism}  & \textbf{business} & \textbf{conspiracy} & \textbf{science} & \textbf{Christianity} & \textbf{lgbt}  \\
  \hline
Anarcho\_Capitalism   & Bitcoin        &  Republican       & Christianity          & DebateAChristian &  feminisms       \\
Bitcoin               & economy        &  conspiritard     & relationship\_advice  & DebateReligion   &  Equality      \\
ronpaul               & entertainment  &  ronpaul          & worldpolitics         & science          &  SubredditDrama          \\
Conservative          & TrueReddit     &  collapse         & MensRights            & videos           &  TwoXChromosomes   \\
Android               & socialism      &  guns             & IAmA                  & news             &  MensRights     \\
ukpolitics            & bestof         &  worldpolitics    & TwoXChromosomes       & Libertarianism   &  offbeat          \\
Equality              & philosophy     &  occupywallstreet & WTF                   & atheism          &  fffffffuuuuuuuuuuuu    \\
  \end{tabular}
\caption{Clusters of subreddits that share dogmatic users. For example, users who are dogmatic on the \textit{conspiracy} subreddit (a place to discuss conspiracy theories) are also likely to be dogmatic on \textit{guns} or \textit{occupywallstreet}.}
  \label{tbl:clusters}
\end{table*}

\subsection{How do dogmatic beliefs cluster? (R2)}

Dogmatism is widely considered to be a domain-specific attitude (for example, oriented towards religion or politics) as opposed to a deeper personality trait \cite{dogmatism}. Here we use Reddit as a lens to examine this idea more closely. Are users who are dogmatic about one topic likely to be dogmatic about others? Do clusters of dogmatism exist around particular topics? To find out, we examine the relationships between subreddits over which individual users are dogmatic. For example, if many users often post dogmatic comments on both the \textit{politics} and \textit{Christianity} subreddits, but less often on \textit{worldnews}, that would suggest \textit{politics} and \textit{Christianity} are linked per a boost in likelihood of individuals being dogmatic in both.

We sample 1000 Reddit users who posted at least once a year between 2007 and 2015 to construct a corpus of 10 million posts that constitute their entire post history. We then annotate these posts using the classifier and compute the average dogmatism score per subreddit per user. For example, one user might have an average dogmatism level of 0.55 for the \textit{politics} subreddit and 0.45 for the \textit{economics} subreddit. Most users do not post in all subreddits, so we track only subreddits for which a user had posted at least 10 times. Any subreddits with an average dogmatism score higher than 0.50 we consider to be a user's dogmatic subreddits. We then count all pairs of these dogmatic subreddits. For example, 45 users have \textit{politics} and \textit{technology} among their dogmatic subreddits, so we consider \textit{politics} and \textit{technology} as linked 45 times. We compute the mutual information \cite{mi} between these links, which gives us a measure of the subreddits that are most related through dogmatism.

We present the results of this analysis in Table \ref{tbl:clusters}, choosing clusters that represent a diverse set of topics. For example, \textit{Libertarianism} is linked through dogmatism to other political communities like \textit{Anarcho\_Capitalism}, \textit{ronpaul}, or \textit{ukpolitics}, as well as other topical subreddits like \textit{guns} or \textit{economy}. Similarly, people who are dogmatic in the \textit{business} subreddit also tend to be dogmatic in subreddits for \textit{Bitcoin}, \textit{socialism}, and \textit{technology}. Notably, when we apply the same mutual information analysis to links defined by subreddits posted in by the same user, we see dramatically different results. For example, the subreddits most linked to \textit{science} through user posts are \textit{UpliftingNews}, \textit{photoshopbattles}, and \textit{firstworldanarchist}, and \textit{millionairemakers}.

Finally, we see less obvious connections between subreddits that suggest some people may be dogmatic by nature. For example, among the users who are dogmatic on \textit{politics}, they are also disproportionately dogmatic on unrelated subreddits such as \textit{science} ($p<0.001$), \textit{technology} ($p<0.001$), \textit{IAmA} ($p<0.001$), and \textit{AskReddit} ($p<0.05$), with p-values computed under a binomial test.

\begin{table}[tb]
\renewcommand{\arraystretch}{1.2}
  \scriptsize
  \begin{tabular}{p{20em}@{\hspace{1em}}p{5em}}
  \textbf{Feature} & \textbf{Direction}  \\
  \hline
total user posts & $\uparrow$ \\
proportion of posts in most active subreddit & $\uparrow$\\
number of subreddits posted in & $\downarrow$\\
average number of posts in active articles & $\downarrow$\\
\end{tabular}
\caption{User behavioral features that are positively and negatively associated with dogmatism. $\uparrow$ means the feature is positively predictive with dogmatism, and $\downarrow$ means the feature is negatively predictive. For example, the more subreddits a user posts in, the less likely they are to be dogmatic. All features are statistically significant ($p<0.001$). }
  \label{tbl:behavior}
\end{table}
\normalsize

\subsection{What user behaviors are predictive of dogmatism? (R3)}

We have shown dogmatism is captured by many linguistic features, but can we discover other high-level user behaviors that are similarly predictive?

To find out, we compute metrics of user behavior using the data sample of 1000 users and 10 million posts described in Section 5.2. Specifically, we calculate (1) \textit{activity}: a user's total number of posts, (2) \textit{breadth}: the number of subreddits a user has posted in, (3) \textit{focus}: the proportion of a user's posts that appear in the subreddit where they are most active, and (4) \textit{engagement}: the average number of posts a user contributes to each discussion they engage in. We then fit these behavioral features to a linear regression model where we predict each user's average dogmatism level. Positive coefficients in this model are positively predictive of dogmatism, while negative coefficients are negatively predictive. %

We find this model is significantly predicitive of dogmatism ($R^2=0.1$, $p<0.001$), with all features reaching statistical significance ($p<0.001$). \textit{Activity} and \textit{focus} are positively associated with dogmatism, while \textit{breadth} and \textit{engagement} are negatively associated (Table \ref{tbl:behavior}). Together, these results suggest dogmatic users tend to post frequently and in specific communities, but are not as inclined to continue to engage with a discussion, once it has begun. 

\subsection{How does dogmatism impact a conversation? (R4)}

How does interacting with a dogmatic comment impact a conversation? Are users able to shrug it off? Or do otherwise non-dogmatic users become more dogmatic themselves?

To answer this question, we sample 600,000 conversations triples from Reddit. These conversations consist of two people (A and B) talking, with the structure: A1 $\rightarrow$ B $\rightarrow$ A2. This allows us to measure the impact of B's dogmatism on A's response, while also controlling for the dogmatism level initially set by A. Concretely, we model the impact of dogmatism on these conversations through a linear regression. This model takes two features, the dogmatism levels of A1 and B, and predicts the dogmatism response of A2. If B's dogmatism has no effect on A's response, the coefficient that corresponds to B will not be significant in the model. Alternatively, if B's dogmatism does have some effect, it will be captured by the model's coefficient.

We find the coefficient of the B feature in the model is positively associated with dogmatism ($p<0.001$). In other words, engagement with a dogmatic comment tends to make a user more dogmatic themselves. This effect holds when we run the same model on data subsets consisting only of dogmatic or non-dogmatic users, and also when we conservatively remove all words used by B from A's response (i.e., controlling for quoting effects).

\section{Related Work}
In contrast to the computational models we have presented, dogmatism is usually measured in psychology through survey scales, in which study participants answer questions designed to reveal underlying personality attributes \cite{dogmatism}. Over time, these surveys have been updated \cite{updated-scale} and improved to meet standards of psychometric validity \cite{dogma-certainty-scale}.

These surveys are often used to study the relationship between dogmatism and other psychological phenomena. For example, dogmatic people tend to show an increased tendency for confrontation \cite{dogma-confrontation} or moral conviction and religiosity \cite{moral-dogma}, and less likelihood of cognitive flexibility \cite{dogma-flexibility}, even among stereotypically non-dogmatic groups like atheists \cite{atheist-dogma}. From a behavioral standpoint, dogmatic people solve problems differently, spending less time framing a problem and expressing more certainty in their solution \cite{problem-solving}. Here we similarly examine how user behaviors on Reddit relate to a language model of dogmatism.

Ertel sought to capture dogmatism linguistically, though a small lexicon of words that correspond with high-level concepts like certainty and compromise \shortcite{dota}. McKenny then used this dictionary to relate dogmatism to argument quality in student essays \shortcite{dogmatism-essays}. Our work expands on this approach, applying supervised models based on a broader set of linguistic categories to identify dogmatism in text.

Other researchers have studied topics similar to dogmatism, such as signals of cognitive style in right-wing political thought \cite{right-wing-cognitive}, the language used by trolls on social media \cite{trolls}, or what makes for impartial language on twitter \cite{impartiality-twitter}. A similar flavor of work has examined linguistic models that capture politeness \cite{politeness}, deception \cite{deception2}, and authority \cite{enron}. We took inspiration from these models when constructing the feature sets in our work. 

Finally, while we examine what makes an opinion dogmatic, other work has pushed further into the structure of arguments, for example classifying their justifications \cite{classify-reason}, or what makes an argument likely to win \cite{winning-arguments}. Our model may allow future researchers to probe these questions more deeply.


\section{Conclusion}
We have constructed the first corpus of social media posts annotated with dogmatism scores, allowing us to explore linguistic features of dogmatism and build a predictive model that analyzes new content. We apply this model to Reddit, where we discover behavioral predictors of dogmatism and topical patterns in the comments of dogmatic users. 

Could we use this computational model to help users shed their dogmatic beliefs? Looking forward, our work makes possible new avenues for encouraging pro-social behavior in online communities. 

\bibliography{emnlp2016}
\bibliographystyle{emnlp2016}

\end{document}